\documentclass[letterpaper, 10 pt, conference]{ieeeconf}
\IEEEoverridecommandlockouts
\usepackage{amsmath}
\usepackage{amssymb}
\usepackage{mathptmx}
\usepackage{graphicx}
\usepackage{multicol}
\usepackage{times}
\usepackage[bookmarks=true]{hyperref}
\hypersetup{
    colorlinks=true,
    urlcolor=magenta,
}
\usepackage{algpseudocode}
\usepackage{algorithm}
\usepackage{mathtools}
\usepackage{threeparttable}
\usepackage[capitalize]{cleveref} 
\usepackage[nolist]{acronym}
\usepackage{siunitx}

\usepackage[final]{changes} 

\newcommand{\FindAnything}{\emph{FindAnything}\xspace}
\newcommand{\FindAnythingMonolithic}{\emph{FindAnything-Monolithic}\xspace}

\usepackage{xspace}

\newcommand{\okvis}{\emph{OKVIS2}\xspace}

\newcommand{\supereight}{\emph{supereight2}\xspace}
\newcommand{\Supereight}{\emph{Supereight2}\xspace}

\usepackage{xifthen}
\usepackage{xparse}
\usepackage{leftidx}
\usepackage{bm}
\usepackage{etoolbox}
\usepackage{amsmath}
\usepackage{multirow, makecell}
\usepackage{xspace}

\newcommand{\bbm}{\begin{bmatrix}}
\newcommand{\ebm}{\end{bmatrix}}

\DeclareMathAlphabet{\mybf}{OT1}{ptm}{b}{n} 
\newcommand{\mybs}[1]{{\bm{#1}}} 
\DeclareMathAlphabet{\mybfi}{OML}{cmm}{b}{it}

\newcommand{\mbf}[1]{
\ifcat\noexpand#1\relax 
\mybs{#1}
\else
\mybf{#1}
\fi
}

\newcommand{\mbfbar}[1]{{\overline{\mbf{#1}}}}
\newcommand{\mbfhat}[1]{{\hat{\mbf{#1}}}}
\newcommand{\mbftilde}[1]{{\tilde{\mbf{#1}}}}

\newcommand{\mbfdot}[1]{{\dot {\mbf{#1}}}}

\NewDocumentCommand{\mbfidentity}{o}{\IfValueTF{#1}{\mbf{I}_{#1\hspace{\rightshift}}}{\mbf{I}}}
\NewDocumentCommand{\mbfzero}{oo}{\IfValueTF{#1}{\mbf{0}_{#1\times#2\hspace{\rightshift}}}{\mbf{0}}}

\newcommand{\cframe}[1]{{\smash{\protect\underrightarrow{\mathcal{F}}_{#1}}}}

\newcommand{\homo}[1]{{\mybfi{#1}}}

\newcommand{\mbfh}[1]{{\homo{#1}}}

\newlength{\leftshift}
\newlength{\rightshift}
\NewDocumentCommand{\vel}{moo}{
	\IfValueTF{#1}{\leftidx{_{#1}}}{}{\mbf v}{\IfValueTF{#2}{_{#2#3\hspace{\rightshift}}}{}}}
\NewDocumentCommand{\veltilde}{moo}{
	\IfValueTF{#1}{\leftidx{_{#1}}}{}{\mbftilde v}{\IfValueTF{#2}{_{#2#3\hspace{\rightshift}}}{}}}
\NewDocumentCommand{\velbar}{moo}{
	\IfValueTF{#1}{\leftidx{_{#1}}}{}{\mbfbar v}{\IfValueTF{#2}{_{#2#3\hspace{\rightshift}}}{}}}
\NewDocumentCommand{\velhat}{moo}{
	\IfValueTF{#1}{\leftidx{_{#1}}}{}{\mbfhat v}{\IfValueTF{#2}{_{#2#3\hspace{\rightshift}}}{}}}
\NewDocumentCommand{\veldot}{moo}{
	\IfValueTF{#1}{\leftidx{_{#1}}}{}{\mbfdot v}{\IfValueTF{#2}{_{#2#3\hspace{\rightshift}}}{}}}

\NewDocumentCommand{\acc}{moo}{
	\IfValueTF{#1}{\leftidx{_{#1}}}{}{\mbf a}{\IfValueTF{#2}{_{#2#3\hspace{\rightshift}}}{}}}
\NewDocumentCommand{\acctilde}{moo}{
	\IfValueTF{#1}{\leftidx{_{#1}}}{}{\mbftilde a}{\IfValueTF{#2}{_{#2#3\hspace{\rightshift}}}{}}}
\NewDocumentCommand{\accbar}{moo}{
	\IfValueTF{#1}{\leftidx{_{#1}}}{}{\mbfbar a}{\IfValueTF{#2}{_{#2#3\hspace{\rightshift}}}{}}}
\NewDocumentCommand{\acchat}{moo}{
	\IfValueTF{#1}{\leftidx{_{#1}}}{}{\mbfhat a}{\IfValueTF{#2}{_{#2#3\hspace{\rightshift}}}{}}}
\NewDocumentCommand{\accdot}{moo}{
	\IfValueTF{#1}{\leftidx{_{#1}}}{}{\mbfdot a}{\IfValueTF{#2}{_{#2#3\hspace{\rightshift}}}{}}}

\NewDocumentCommand{\rotvel}{moo}{
	\IfValueTF{#1}{\leftidx{_{#1}}}{}{\mbf $\omega$}{\IfValueTF{#2}{_{#2#3\hspace{\rightshift}}}{}}}
\NewDocumentCommand{\rotveltilde}{moo}{
	\IfValueTF{#1}{\leftidx{_{#1}}}{}{\mbftilde $\omega$}{\IfValueTF{#2}{_{#2#3\hspace{\rightshift}}}{}}}
\NewDocumentCommand{\rotvelbar}{moo}{
	\IfValueTF{#1}{\leftidx{_{#1}}}{}{\mbfbar $\omega$}{\IfValueTF{#2}{_{#2#3\hspace{\rightshift}}}{}}}
\NewDocumentCommand{\rotvelhat}{moo}{
	\IfValueTF{#1}{\leftidx{_{#1}}}{}{\mbfhat $\omega$}{\IfValueTF{#2}{_{#2#3\hspace{\rightshift}}}{}}}
\NewDocumentCommand{\rotveldot}{moo}{
	\IfValueTF{#1}{\leftidx{_{#1}}}{}{\mbfdot $\omega$}{\IfValueTF{#2}{_{#2#3\hspace{\rightshift}}}{}}}

\newcommand{\T}[2]{{\mbfh T}{_{#1#2\hspace{\rightshift}}}} 

\newcommand{\N}{\mathbb{N}}
\newcommand{\R}{\mathbb{R}}

\newcommand{\Title}{FindAnything: Open-Vocabulary and Object-Centric Mapping\\
for Robot Exploration in Any Environment}
\title{\Title}

\author{Sebasti\'an Barbas Laina$^{1,5,*}$, Simon Boche$^{1,*}$, Sotiris Papatheodorou$^{1,3,4,5,6,*}$,\\ Simon Schaefer$^{1}$, Jaehyung Jung$^{1}$, Helen Oleynikova$^{2}$, Stefan Leutenegger$^{1,2,3,5,6}$
\thanks{This work was supported by the Technical University of Munich, MIRMI, a donation by Google, the State of Bavaria through the REACT project, the TUM Innovation Network CoConstruct, ETH Zurich, and the EU Horizon projects DigiForest \scriptsize{(101070405)} and AUTOASSESS \scriptsize{(101120732)}.}%
\thanks{$^{1}$Mobile Robotics Lab, School of Computation, Information and Technology, Technical University of Munich. E-mail addresses: \texttt{\{sebastian.barbas, simon.boche, sotiris.papatheodorou, simon.k.schaefer, jaehyung.jung, stefan.leutenegger\}@tum.de}}%
\thanks{$^{2}$Mobile Robotics Lab, Department of Mechanical and Process Engineering, ETH Z{\"u}rich. E-mail address: \texttt{\{lestefan, oelena\}@ethz.ch}}%
\thanks{$^{3}$Department of Computing, Imperial College London. E-mail address: \texttt{s.leutenegger@ic.ac.uk}}%
\thanks{$^{4}$Department of Electrical and Computer Engineering, University of Patras. E-mail address: \texttt{s.papatheodorou@ac.upatras.gr}}%
\thanks{$^{5}$Munich Institute of Robotics and Machine Intelligence (MIRMI).}%
\thanks{$^{6}$Munich Center for Machine Learning (MCML).}%
\thanks{$^{*}$ Equal contribution.}%
}

\usepackage{arydshln} 

\begin{document}
\maketitle
\thispagestyle{empty}
\pagestyle{empty}

\begin{acronym}
\acro{ATE}{Absolute Trajectory Error}
\acro{CNN}{Convolutional Neural Network}
\acro{CPU}{Central Processing Unit}
\acro{CoM}{Centre of Mass}
\acro{DoF}{Degree of Freedom}
\acro{EKF}{Extended Kalman Filter}
\acro{ESDF}{Euclidean Signed Distance Field}
\acro{FoV}{Field of View}
\acro{GNN}{Graph Neural Network}
\acro{GP-GPU}{General-Purpose computing on \acsp{GPU}}
\acro{GPS}{Global Positioning System}
\acro{GPU}{Graphics Processing Unit}
\acro{ICP}{Iterative Closest Point}
\acro{IMU}{Inertial Measurement Unit}
\acro{LI-SLAM}{\acs{LiDAR}-Inertial \acs{SLAM}}
\acro{LLM}{Large Language Model}
\acro{LQR}{Linear Quadratic Regulator}
\acro{LVI-SLAM}{\acs{LiDAR}-Visual-Inertial \acs{SLAM}}
\acro{LiDAR}{Light Detection and Ranging}
\acro{MAV}{Micro Aerial Vehicle}
\acro{MLP}{Multi-Layer Perceptron}
\acro{MPC}{Model Predictive Control\acroextra{(ler)}}
\acro{NBV}{Next-Best-View}
\acro{NMPC}{Non-linear Model Predictive Control\acroextra{(ler)}}
\acro{NeRF}{Neural Radiance Field}
\acro{OMPL}{Open Motion Planning Library}
\acro{PID}{Proportional Integral Derivative}
\acro{QP}{Quadratic Programming}
\acro{RAM}{Random Access Memory}
\acro{RGB-D}{\acs{RGB}-Depth}
\acro{RGB}{Red Green Blue}
\acro{RMSE}{Root Mean Squared Error}
\acro{ROS}{Robot Operating System}
\acro{RRT}{Rapidly-exploring Random Tree}
\acro{SDF}{Signed Distance Field}
\acro{SFC}{Safe Flight Corridor}
\acro{SLAM}{Simultaneous Localisation and Mapping}
\acro{TSDF}{Truncated Signed Distance Field}
\acro{VI-SLAM}{Visual-Inertial \acs{SLAM}}
\acro{VIO}{Visual-Inertial Odometry}
\acro{VI}{Visual-Inertial}
\acro{VL}{Vision-Language}
\acro{VLM}{Vision-Language Model}
\acro{VO}{Visual Odometry}
\acro{eSAM}{\textit{EfficientSAM}}
\acro{SnR}[S\&R]{Search and Rescue}
\acro{SAM}{\textit{SegmentAnything}}

\acrodefplural{DoF}{Degrees of Freedom}

\acroindefinite{ESDF}{an}{a}
\acroindefinite{FoV}{an}{a}
\acroindefinite{LI-SLAM}{an}{a}
\acroindefinite{LLM}{an}{a}
\acroindefinite{LQR}{an}{a}
\acroindefinite{LVI-SLAM}{an}{a}
\acroindefinite{MAV}{an}{a}
\acroindefinite{MLP}{an}{a}
\acroindefinite{MPC}{an}{a}
\acroindefinite{NBV}{an}{a}
\acroindefinite{NMPC}{an}{a}
\acroindefinite{RGB-D}{an}{a}
\acroindefinite{RGB}{an}{a}
\acroindefinite{RMSE}{an}{a}
\acroindefinite{RRT}{an}{a}
\end{acronym}

\begin{abstract}

Geometrically accurate and semantically expressive map representations have proven invaluable for robot deployment and task planning in unknown environments. Nevertheless, real-time, open-vocabulary semantic understanding of large-scale unknown environments still presents open challenges, mainly due to computational requirements. 
In this paper we present \FindAnything, an open-world mapping framework that incorporates vision-language information into dense volumetric submaps. Thanks to the use of vision-language features, \FindAnything combines pure geometric and open-vocabulary semantic information for a higher level of understanding. It proposes an efficient storage of open-vocabulary information through the aggregation of features at the object level. Pixel-wise vision-language features are aggregated based on eSAM segments, which are in turn integrated into object-centric volumetric submaps, providing a mapping from open-vocabulary queries to 3D geometry that is scalable also in terms of memory usage.
We demonstrate that \FindAnything performs on par with the state-of-the-art in terms of semantic accuracy while being substantially faster and more memory-efficient, allowing its deployment in large-scale environments and on resource-constrained devices, such as \acsp{MAV}. 
We show that the real-time capabilities of \FindAnything make it useful for downstream tasks, such as autonomous \acs{MAV} exploration in a simulated Search and Rescue scenario. \\ 
Project Page: \textmd{\url{https://ethz-mrl.github.io/findanything/}}.

\end{abstract}

\IEEEpeerreviewmaketitle


\section{Introduction}
One key application of robotics is \ac{SnR} and disaster response: where robots should take the place of humans in hazardous or inaccessible environments, and provide useful, safety-critical information to first responders. \Acp{MAV} are particularly relevant here, as they move freely in 3D space and can easily access areas out of reach of humans and ground-based robots.

A robot that is useful for emergency situations requires safe operation with minimal human supervision, while giving remote operators live, up-to-date information about its environment: not just geometry and appearance, but also higher-level semantic information and object locations. 

We propose \FindAnything, a mapping system that runs online on-board \iac{MAV} to address these requirements: creating a scalable, submap-based volumetric map that includes the scene geometry, appearance, and also encodes \textit{open-vocabulary} features in an object-centric way, allowing operators to query the map live for locations of rooms, objects, or properties of the scene.

\begin{figure}[t]
    \centering
    \includegraphics[width=\columnwidth]{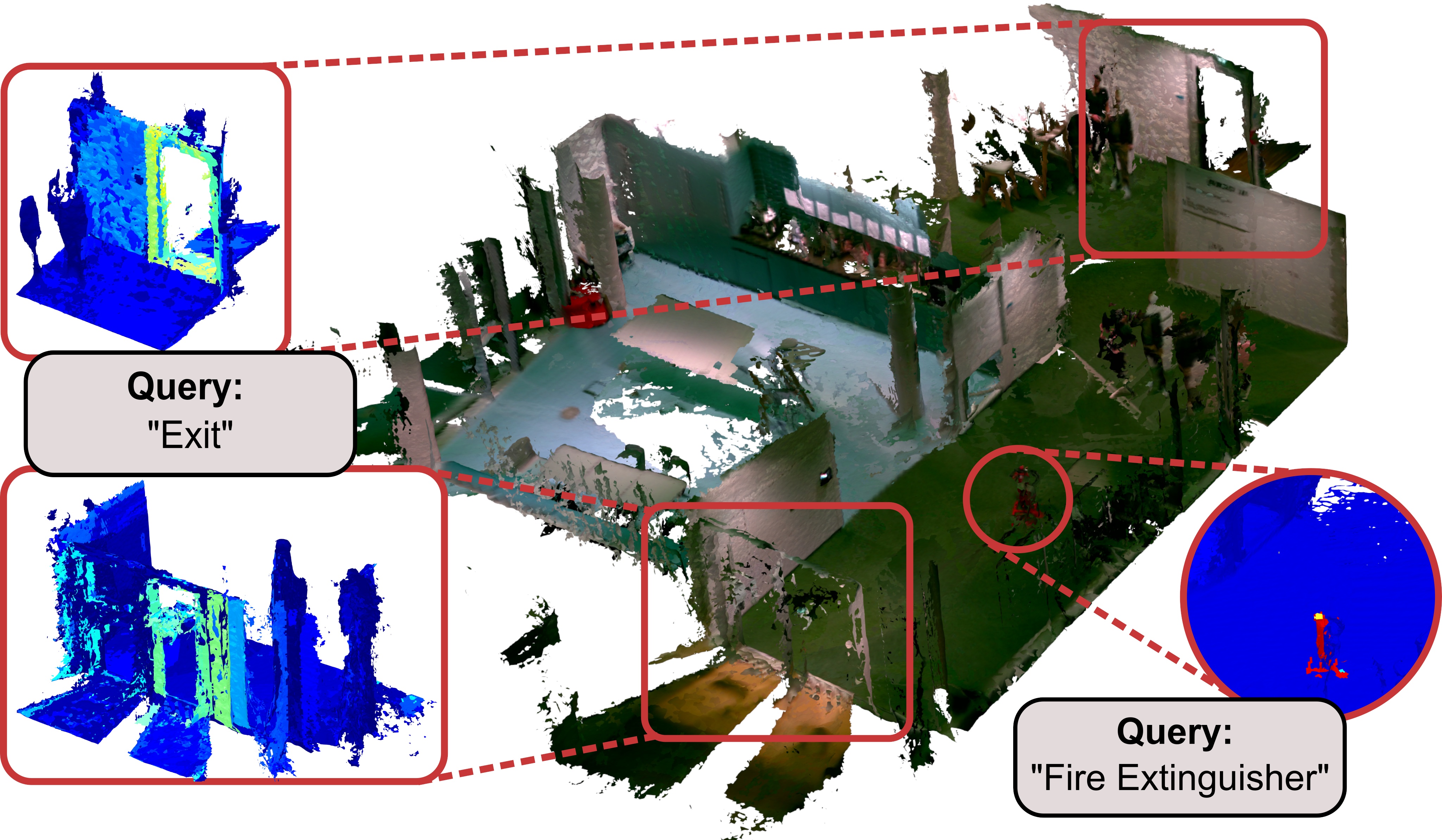}
    \caption{Real-world \acs{MAV} exploration demonstration of \FindAnything in an office environment, including a kitchen.
    Colored meshes extracted from volumetric maps are shown alongside the 3D CLIP activations in our object-centric map representation for the language queries \textit{``fire extinguisher''} and \textit{``exit''}, which are potential points of interest in a fire scenario.}
    \label{fig:real-world}
\end{figure}

For instance, consider a fire emergency, where a robot might be assigned to locate objects or tools useful for firefighting, e.g.\ fire extinguishers, or the nearest exit (see~\cref{fig:real-world}).
To succeed in this demanding and highly time-sensitive task, the robot needs the capability to incrementally reconstruct the unknown scene online and in real-time while safely navigating through it.
To ensure scalability to large environments (e.g.\ multistory houses), the underlying map representation must be memory-efficient for deployment on resource-constrained platforms, such as \acp{MAV}.
Finally, the described task also requires advanced scene understanding capabilities that go beyond purely geometric reasoning but also incorporate generalizable semantic reasoning without relying on prior knowledge of the scene.

Volumetric maps are a common scene representation for \acp{MAV}, as they are useful for both on-board navigation and as a map interpretable by robot operators~\cite{AnnaDai,oleynikova2020open}.
To go beyond mere geometry and appearance and enable a higher level of scene understanding, many approaches have been studied to integrate class-level semantic information into volumetric maps~\cite{mccormac2017semanticfusion, grinvald2019volumetric, rosinol2021kimera}. However, these methods require an a priori known set of classes, making the semantic information compact to store (as the number of classes is finite), but limiting the expressiveness and applicability of the maps.

\Ac{VL} models such as CLIP~\cite{clip} do not have these restrictions and instead store semantic information as a high-dimensional \textit{feature embedding} that can be queried in real time for matching a human language description.
While this gives such models incredible flexibility, it also comes with a computational and memory cost: where a class label can be stored as a single integer, feature embeddings are generally hundreds of floating-point values.
This makes them challenging to aggregate into a 3D map on resource-constrained hardware.
To address this, recent works have explored more compact feature representations~\cite{alama2025rayfronts} or combined \ac{VL} and segmentation foundation models~\cite{clio,werby2024hierarchical,j2023conceptfusion}, such as \ac{SAM}~\cite{sam} or more efficient variants of it, e.g.\ \ac{eSAM}~\cite{xiong2024efficientsam}.
However, most of these approaches still fall short of the computational requirements to build large-scale maps online and on-board \iac{MAV}.

We propose \FindAnything, a system for large-scale volumetric mapping with open-vocabulary semantic capabilities.
Dividing the volume into smaller submaps allows for drift-correcting mechanisms such as loop closures, enabling the deployment in large-scale environments where some degree of state estimation drift is inevitable.
Our method aggregates open-vocabulary information over time at the level of objects or object parts.
Oversegmentation of objects into smaller entities allows for fine-grained queries, while the generalization capabilities of \ac{VL} features preserve an understanding of larger objects or concepts.
The object-centric map representation significantly reduces the memory requirements while allowing the use of foundation models that trade accuracy for speed.
We show that these characteristics make \FindAnything suitable for downstream tasks, such as the deployment of \iac{MAV} for autonomous exploration in a simulated fire emergency. 
The main contributions of this paper are:
\begin{itemize}
    \item A method to aggregate high-dimensional \ac{VL} features into a volumetric map in a memory-efficient, object-centric way: using image-based semantic oversegmentation, segment tracking and association, and feature embedding merging.
    \item An integration of the proposed object-centric \ac{VL} feature mapping approach with a submapping-based visual(-inertial) SLAM system, enabling large-scale, online, compute- and memory-efficient mapping even on resource-constrained platforms.
    \item Evaluation in simulation and real-world benchmarks showing that \FindAnything achieves semantic accuracy competitive to the state-of-the-art while requiring shorter computational times and up to 60\% less memory.
    \item A showcase of a downstream exploration task on-board an MAV, demonstrating that \FindAnything can be used to help guide robot exploration using natural language.
\end{itemize}

\section{Related Work}
\label{sec:related-work}
\subsection{Foundation Models}
Foundation models, trained on internet-scale data, are a cornerstone for tackling a wide range of tasks in a zero-shot manner.
Models like CLIP~\cite{clip}, LLaVA~\cite{liu2023llava}, and LSeg~\cite{li2022lseg} enable reasoning about images in natural language by encoding visual and textual information in a single feature space.

While these models were trained using contrastive learning on image patches~\cite{clip}, significant advancements have been made in precisely localizing features in images~\cite{ghiasi2022sovis, zhong2022regionclip, li2022lseg}.
However, challenges remain in accurately representing complex object shapes and long-tailed object categories~\cite{j2023conceptfusion}. 

In this work, we address these limitations by integrating the \ac{VL} foundation model CLIP~\cite{clip} with the general-purpose segmentation model \ac{eSAM}~\cite{xiong2024efficientsam}.
By accumulating extracted features at the object (or part) level within our probabilistic mapping framework, we effectively harness the extensive open-vocabulary capabilities of the \ac{VL} model without relying on its ability to produce highly precise feature maps.
This approach allows for robust semantic understanding, even for complex and diverse object types.

\begin{figure*}[t]
    \centering
    \includegraphics[width=0.9\textwidth]{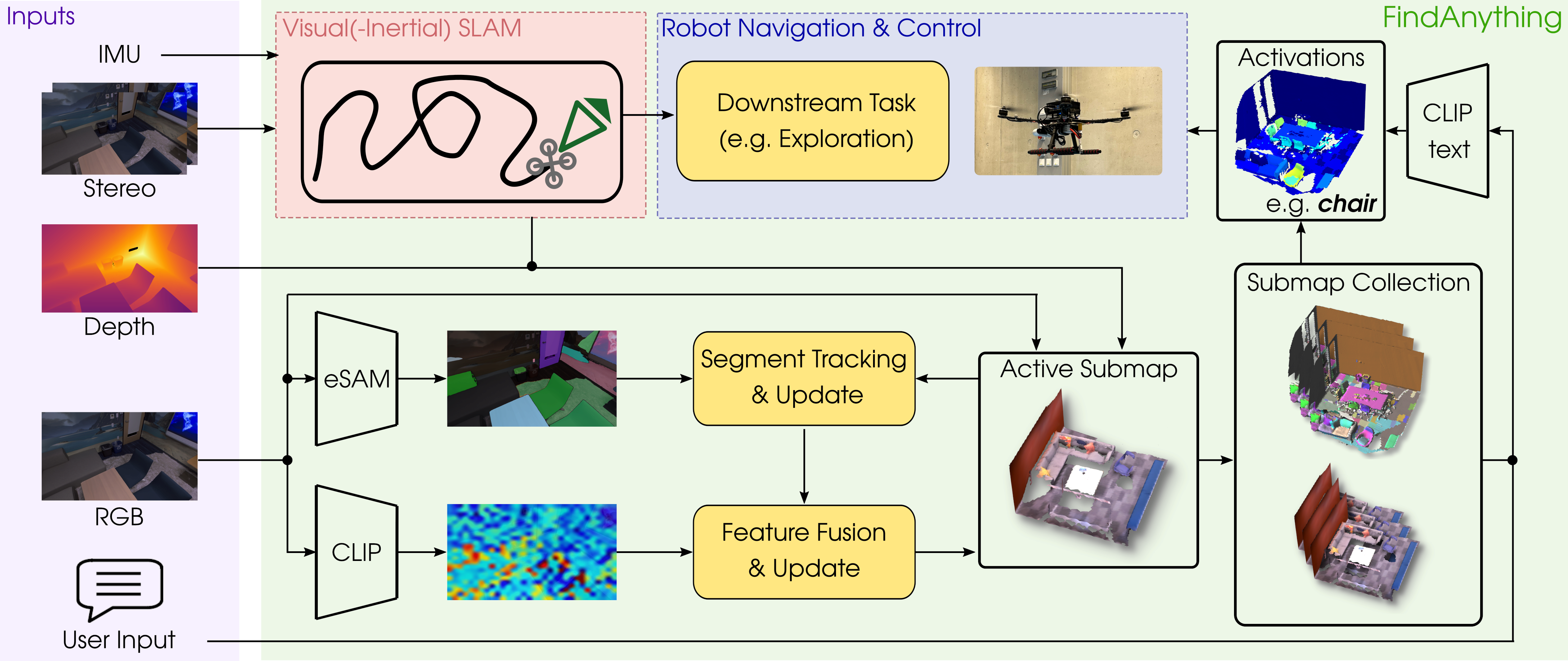}
    \caption{Overview of \FindAnything: \acs{VI-SLAM} provides estimated poses to integrate depth and RGB images into volumetric occupancy submaps. \acs{eSAM}~\cite{xiong2024efficientsam} creates object proposals from the RGB images which are tracked against the current map. CLIP~\cite{clip} features are aggregated per object mask and fused into the current submap. The submaps can be directly used in downstream tasks, such as autonomous exploration. In addition, natural language user input is translated into a CLIP feature to find regions of high activation in the map and guide the exploration towards those areas of interest.}
    \label{fig:system-overview}
\end{figure*}

\subsection{Open-Vocabulary-based Mapping \& Navigation}
Recent works have explored extending 2D foundation models into 3D without relying on internet-scale 3D datasets.
Some methods embed \ac{VL} features into NeRFs~\cite{lerf2023,liao2024ovnerf} or Gaussian splats~\cite{qin2023langsplat}.
While these approaches generate photorealistic renders of the environment, they require time-consuming training before deployment, limiting their scalability to large environments or their use onboard robots.

To address this limitation, \cite{yamazaki2024open} proposed a method to fuse \ac{VL} features into a monolithic TSDF map, allowing for efficient updates and queries but remaining constrained in map size due to the memory requirements.
ConceptFusion~\cite{j2023conceptfusion} and related works~\cite{werby2024hierarchical,lu2023ovir} directly store features in a point cloud, combining region-level and global embeddings. \cite{wei2024ovexp,huang23vlmaps} instead integrate features into a 2D top-down map. More recently, several methods~\cite{clio, werby2024hierarchical, martins2024ovo, schmid24khronos, wang2024octree} focus on accumulating \ac{VL} features at the instance level using 3D scene graphs or instance-level point clouds.
While these approaches offer improved scalability, they either do not preserve full 3D geometric information, essential for downstream tasks, or are prone to map errors as they do not allow for drift-correcting mechanisms such as loop closures. 

In contrast, RayFronts~\cite{alama2025rayfronts} proposes an open-set semantic 3D mapping framework for outdoors robotic exploration.
To enhance efficiency and semantic accuracy, they propose a fine-grained dense vision-language encoder based on RADIO~\cite{ranzinger2024radio}. Nevertheless, fusing \ac{VL} features at the voxel level into a VDB-based occupancy map results in high memory usage, limiting its applicability to larger environments.

\FindAnything aims to combine the best of both worlds. Similar to~\cite{clio}, we use segmentation foundation models. However, instead of scene graphs or point clouds we utilize instance-level volumetric occupancy maps with adaptive resolution. Also, instead of tracking in 3D, our segment tracking is done in image space by projecting map objects into the image plane. This alleviates volumetric discrepancies between an object and its current view. This approach enables accumulating \ac{VL} features in real-time, efficient natural language queries, and error corrections by coupling mapping with the \ac{SLAM} pipeline. Aggregating features at the instance-level significantly reduces memory usage and improves scalability.
Unlike prior works, we show that our system runs completely online, even on resource-constrained devices. 
By preserving detailed 3D geometric as well as object-level semantic information, our method overcomes the limitations of prior works and is well-suited for large-scale environments, and applications such as exploration.

\section{Methodology}
\label{section:approach}

In this section, we present the modules that form our system, a schematic overview of which is presented in \cref{fig:system-overview}.

\subsection{Notation and Definitions}
\label{section:notation}
The used \ac{VI-SLAM} system tracks a moving body with a mounted IMU and several cameras relative to a static world coordinate frame $\cframe{W}$. The robot pose, expressed in the IMU frame $\cframe{S}$, is denoted as $\T{W}{S}$. Images are presented in matrix form, $\textbf{I}$, and we obtain values at a pixel coordinate $\textbf{u}$ (or pixel range as mask) via $\textbf{I}[\textbf{u}]$.

\subsection{VI-SLAM}
\label{sec:slam}

Our state estimation module is based on the multi-sensor \ac{SLAM} system OKVIS2-X~\cite{boche2025okvis2x}, an extension of \okvis \cite{okvis2}, that incorporates depth information to improve the state-estimation accuracy. \ac{SLAM} poses are used to integrate depth information into volumetric occupancy submaps. A new state $\mbf{x}_{l}$ is estimated for a new pair of stereo images at timestamp $l$. The state-estimator performs real-time sliding window and loop closure optimization. As a result, for a new pair of frames, the state $\mbf{x}_l$ is estimated and prior states $\mbf{x}_{l-k}$ are updated as necessary, including loop closures.

\subsection{Volumetric Occupancy Mapping}
\label{sec:VolMap}
The geometric information is represented by partitioning the environment into volumetric submaps using the \supereight~\cite{supereight2} mapping framework. \Supereight's occupancy map representation allows differentiating between occupied, free and unmapped regions, making it directly usable for safe path planning and navigation. By employing submaps, we achieve scalability to large environments while using shallower data structures for mapping, enabling faster data access. We track the rigid transformations between submaps and update them after each SLAM optimization, including loop closures, following the scheme proposed in~\cite{laina2024scalable}, where each submap is associated to a keyframe.


\subsection{Vision-Language Feature Fusion}
\label{section:language_fusion}

As \iac{VL} feature extractor, we use CLIP~\cite{clip} ViT-L/14 pre-trained for images of $336 \times 336$ pixels. Following~\cite{zhou2022extract}, we obtain an image $\textbf{F}$ with a 768-dimensional feature embedding per pixel. Aggregating these features into volumetric submaps at a voxel level would lead to vast memory requirements, limiting the scalability of the system to larger environments. Instead, we use an object-centric approach, in which segments obtained from image-based segmentation are associated to voxels. For subsequent frames, we perform an image-to-map segment tracking. This choice allows us to decouple the voxel resolution from the language representation, enabling mapping at high resolutions without significantly increasing memory usage, a limiting factor on resource-constrained robots.

We use the \ac{eSAM}~\cite{xiong2024efficientsam} foundation model, a lightweight SAM \cite{sam} version with faster inference time, to obtain a collection of binary segment masks $\mathcal{E}_t \in \{\textbf{E}_0,\textbf{E}_1, ...,\textbf{E}_m\}$.
As the agent moves through the environment, we track or split the objects that have been previously added to the submap while adding new segments to previously unsegmented regions.

To construct an object-centric map representation, we extend \supereight by incorporating segment IDs into occupied voxels.
Each segment ID, referred to as $k$, stores an associated average language feature $\mbf{\bar{f}}_k$ encoding open-vocabulary semantic information, and the total number of pixels $N_{k}$ that have been used to update this segment.
To update the segment IDs of a submap at time step $t$, we require a depth image $\mbf D_t$, the robot pose at that time $\T{W}{S_t}$ and a corresponding image with per-pixel segment IDs $\textbf{K}_t$.
To track objects, we also render a collection of binary segment masks and associated segment IDs, $\mathcal{R}_t \in \{(k_0, \textbf{R}_0), (k_1, \textbf{R}_1), ..., (k_p, \textbf{R}_p)\}$, of the current submap into the image frame of given pose $\T{W}{S_t}$. 

\begin{figure}[t]
    \centering
    \includegraphics[width=0.95\columnwidth]{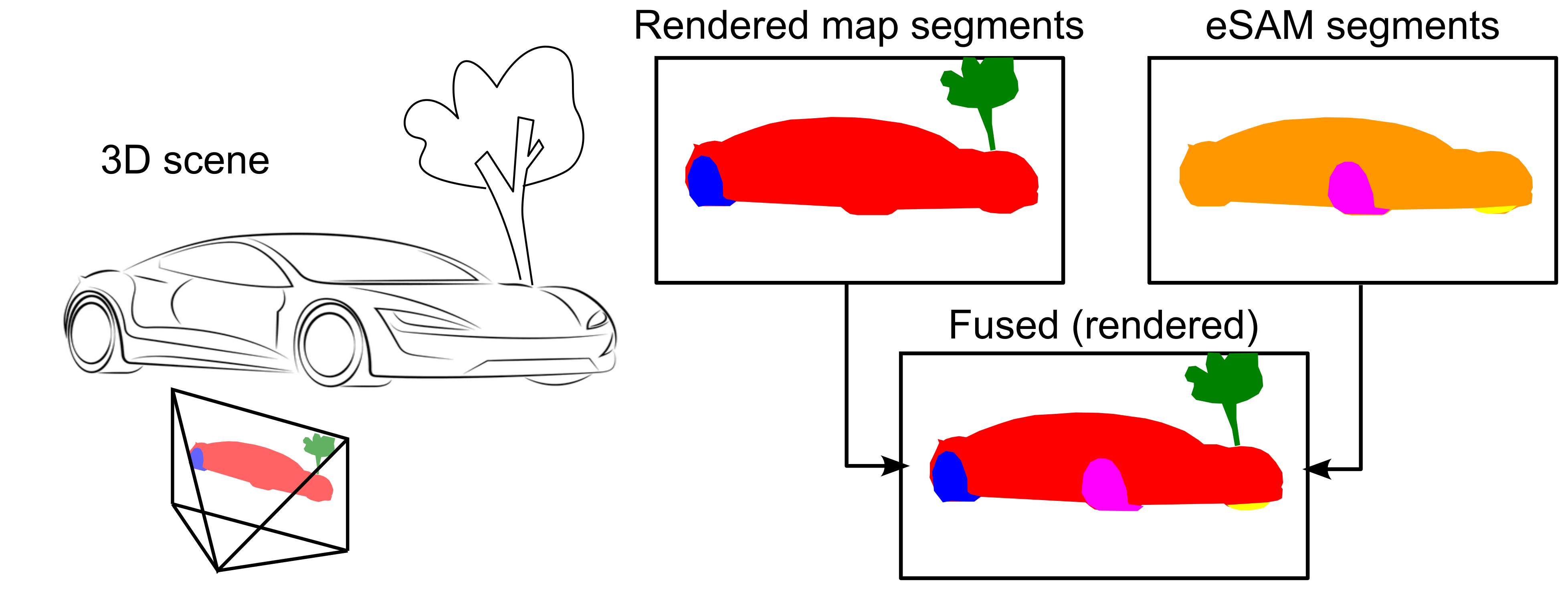}
    \caption{Schematic of our as-fine-as-possible segmentation strategy. A segment image of the map is rendered from the current camera pose. Rendered segments and eSAM segment proposals are fused into the map favoring the smallest option available.}
    \label{fig:oversegment_schematic}
    \vspace{-0.3cm}
\end{figure}

We follow an as-fine-as-possible segmentation strategy to represent the objects in the smallest partition proposed by \ac{eSAM} over time. Segment tracking and oversegmentation is based on 2D image overlap between \ac{eSAM} segments $\mathcal{E}_t$ and rendered segments $\mathcal{R}_t$ from the current submap. Our method greedily partitions both segment sources, if possible, by prioritizing smaller segments (above a threshold), producing an image $\textbf{K}_t$ with per-pixel segment IDs to be integrated into the current submap. A visual schematic of this approach is presented in \cref{fig:oversegment_schematic}. \acs{VL} features are then aggregated per segment over all pixels with an ID in $\textbf{K}_t$. This strategy allows for long-term tracking and is fully computed on a CPU, freeing the GPU resources for the foundation models. This oversegmentation of objects into smaller entities allows for fine-grained queries, while the VL features preserve a semantic understanding of larger objects or concepts, allowing to group similar segments upon query time. The assumption of semantic consistency, together with that of spatial consistency between submaps, renders segment tracking across maps unnecessary.

After segment tracking, a weighted average update is performed on the per-segment feature vectors using the per-pixel segment ID image $\textbf{K}_t$ and the CLIP image $\textbf{F}_t$.
For each segment ID $k$ in $\textbf{K}_t$ we compute the set of pixel coordinates $\mathcal{N} = \{\mbf{u}_1, \mbf{u}_2, ..., \mbf{u}_N\}$ where $\textbf{K}_t$ has the value $k$. The \ac{VL} feature is then updated as
\begin{align}
\label{eq:feature_updates}
    \mbf{\bar{f}}_{k} &\leftarrow \frac{N_{k} \mbf{\bar{f}}_{k} + \sum_{i=1}^{N} \textbf{F}_t[\mbf{u}_i]}{N_{k} + N }, \\
    N_{k} &\leftarrow N_{k} + N ,
\end{align}
where $N_{k}$ is the number of pixels previously associated with segment ID $k$, while $N$ is the number of pixels in the tracked segment in the current frame. The average language feature descriptor of segment $k$ is denoted as $\mbf{\bar{f}}_{k}$.
These weighted updates from different viewpoints improve the consistency of the \ac{VL} embedding representation. 
\section{Evaluation and Experiments}
To showcase the accuracy and computational efficiency of our proposed system, we benchmark \FindAnything on a number of standard datasets both in indoor (\cref{sec:results-replica}) and outdoor (\cref{sec:results-kitti}) scenarios. As motivated by our application requirements, we focus on evaluating semantic map accuracy as well as runtime and memory usage.

We additionally showcase the usefulness of \FindAnything in an integrated real-world experiment, where \FindAnything is used as the mapping back-end for online exploration in a simulated firefighting scenario (see \cref{sec:evaluation-realworld}).

\subsection{Evaluation Criteria}
For the semantic accuracy of the aggregated \ac{VL} features, we follow the standard evaluation setup of~\cite{alama2025rayfronts, clio, j2023conceptfusion, gu2024conceptgraphs} and report accuracy as the class-mean recall (mAcc) and the frequency-weighted mean intersection-over-union (f-mIOU) of closed-set predictions of the semantic classes. 

We demonstrate \mbox{\FindAnything's} suitability for online deployment in resource-constrained devices by evaluating runtime of our system against different state-of-the-art open-vocabulary 3D mapping approaches on the Replica dataset~\cite{replica}, a typical indoor dataset, in \cref{sec:results-replica}, and additionally memory usage in SemanticKITTI~\cite{semantic-kitti}, a large-scale outdoor dataset, in \cref{sec:results-kitti}.

As a main baseline, we chose to benchmark against RayFronts~\cite{alama2025rayfronts}, one of the few established approaches for large-scale volumetric open-vocabulary mapping. As RayFronts runs mainly on the GPU, we report memory as the combined GPU and RAM usage. For fair comparisons, we ensured equal voxel resolution and depth range limits and set a batch size of $1$ to mimic online mapping. Furthermore, we disable their ray frontiers representation for mapping beyond depth-sensing range to limit the evaluation to the semantic accuracy and memory usage of the underlying volumetric map.
Results for other papers are mainly taken from the respective papers. All results from competitors were obtained using ground-truth (GT) poses. For \FindAnything, results are shown for both GT and SLAM poses to evaluate both ideal and realistic real-world performance. 

We also evaluate \FindAnything's applicability for downstream tasks such as exploration, by evaluating mesh completion and accuracy of an exploration approach with and without using semantic information in \cref{sec:evaluation-exploration}.

\subsection{Indoor Dataset: Replica}
\label{sec:results-replica}
Our first evaluation seeks to answer multiple questions: 1) how does the accuracy of {\FindAnything} compare to state-of-the-art semantic mapping approaches,
2) what impact does submapping versus monolithic mapping have in small-scale environments,
and 3) how much faster is {\FindAnything} in processing a complete sequence than existing approaches?

To answer these questions, we evaluate on the Replica~\cite{replica} dataset, using  the same pre-rendered sequences as~\cite{j2023conceptfusion}. To use Replica (a monocular dataset) in our stereo SLAM system, we also render a synthetic color image 6 cm to the right of the original one using Habitat-Sim~\cite{habitat19iccv}. In our experiments, the voxel resolution was set to \SI{5}{\centi\meter}.

Results for the semantic accuracy are presented in \cref{tab:replica-semantics}, where we demonstrate that our proposed approach is competitive to the state-of-the-art. In fact, thanks to the modular design of \FindAnything, it can achieve the highest semantic accuracy when using the same VL encoder (NARADIO) as RayFronts.
Despite the higher accuracy of NARADIO, we decided to use CLIP features due to the lower dimensionality of its embeddings, resulting in a lower memory footprint at the expense of minor degradation in semantic accuracy.


As a consequence of the high SLAM accuracy on such small-scale datasets, the difference in semantic accuracy between our method with GT or SLAM poses is negligible.
It is important to note that there is a minor randomness in our results.
Even with GT poses, the submap creation still depends on SLAM running in the background.
Due to the higher global consistency of submapping in the presence of loop closures, there is still a minor improvement over \FindAnythingMonolithic, a version using a monolithic map.


\begin{table}
\caption{3D semantic map accuracy (Replica).} \label{tab:replica-semantics}
\centering
\renewcommand{\arraystretch}{1.1} 
\begin{tabular}{c | l| c c }
    & Method & mAcc & f-mIoU \\
    \hline
    \multirow{11}{*}{\rotatebox[origin=c]{90}{ \scalebox{0.9}{\scriptsize GT} }} & ConceptFusion$^{1}$~\cite{j2023conceptfusion} & 24.16 & 31.31 \\
    & ConceptFusion$^{1}$~\cite{j2023conceptfusion} + SAM & 31.53 & 38.70\\
    & ConceptGraphs$^{1}$~\cite{gu2024conceptgraphs} & 40.63 & 35.95 \\
    & HOV-SG (Vit-H-14)$^{1}$~\cite{werby2024hierarchical} & 30.40 & 38.60 \\
    & Clio-batch$^{1}$~\cite{clio} & 37.95 & 36.98 \\
    & OVO-SLAM$\dag$~\cite{martins2024ovo} & 32.18 & 46.19 \\
    & Octree-Graph (OVSeg)$^{1}$~\cite{wang2024octree} & 41.40 & 55.30 \\ 
    & RayFronts (NACLIP)~\cite{alama2025rayfronts} & 26.44  & 32.66 \\ 
    & RayFronts (NARADIO)~\cite{alama2025rayfronts} & \underline{52.90}  & \underline{64.97} \\
    & \FindAnything$\dag$ & 44.48 & 62.01 \\
    & \FindAnything & 48.87 & 62.71 \\
    & \FindAnything (NARADIO) & \textbf{53.55} & \textbf{66.91} \\
    \hline
    \multirow{2}{*}{\rotatebox[origin=c]{90}{ \hspace{-0.1cm}\scalebox{0.9}{\scriptsize SLAM}}} & \FindAnythingMonolithic & 47.57 & 61.37 \\
    &\FindAnything  & 48.80 & 62.91
\end{tabular}

\begin{tablenotes}
   \item $^{1}$ Results taken from the respective papers. Words in parentheses indicate VL encoders if different versions are available.
   \item $\dag$ Only processing every 10th frame.
\end{tablenotes}
\end{table}

To demonstrate the real-time capabilities of our system, we show the average runtime per sequence in \cref{tab:replica-timings} and provide a detailed time distribution of the different stages of the method in \cref{tab:replica-process-timings}.
As seen in \cref{tab:replica-timings}, our method is substantially faster than \cite{werby2024hierarchical} and faster than other competitors, e.g.\ OVO-SLAM~\cite{martins2024ovo} or RayFronts~\cite{alama2025rayfronts}.
As OVO-SLAM only processes every tenth frame, we evaluate \FindAnything with the same processing strategy and demonstrate that it is faster and semantically more accurate. The minor semantic degradation in this evaluation mainly is due to the limited motion in the dataset, which is not assured in other scenarios.

\begin{table}
\caption{Mean Replica sequence processing time} \label{tab:replica-timings}
\centering
\renewcommand{\arraystretch}{1.1} 
\begin{tabular}{l|c}
    Method & Mean time per sequence \\
    \hline
    OVO-SLAM$^\dag$~\cite{martins2024ovo} & \underline{3m 2s} \\
    \FindAnything$^\dag$ &  \textbf{1m 19s} \\

    \hline
    HOV-SG~\cite{werby2024hierarchical} & 11h 12m \\
    Rayfronts~\cite{alama2025rayfronts} & \underline{9m 19s}  \\
    \FindAnything &  \textbf{5m 24s} \\
\end{tabular}

\begin{tablenotes}
   \item HOV-SG timings taken from~\cite{martins2024ovo}, which used an NVIDIA RTX 3090 GPU. The other timings are obtained using an NVIDIA RTX 4500.
   \item $^\dag$ Only processing every 10th frame.
\end{tablenotes}
\end{table}


\subsection{Large-scale outdoor dataset: Semantic KITTI}
\label{sec:results-kitti}
Our second evaluation mainly focuses on two objectives. First, we evaluate the scalability of \FindAnything to large environments in terms of semantic map accuracy, processing time and memory usage and benchmark it against RayFronts. Second, we justify our system design through a series of ablation studies. These studies specifically investigate: (a) the choice of the segmentation model; (b) the proposed strategy for \ac{VL} feature fusion and oversegmentation; and (c) the partitioning of the map into submaps in a real-world scenario using SLAM poses.
\Cref{tab:semantic-kitti} presents the semantic accuracy as well as the average sequence processing time $T_{\mathrm{avg}}$ and the average per-sequence memory usage $M_{\mathrm{avg}}$ on the SemanticKITTI dataset~\cite{semantic-kitti}, a large-scale outdoor dataset. For semantic evaluation, we exclude classes labeled as ``moving'' or ``other''. Depth images were obtained by pre-processing stereo images with~\cite{wen2025foundationstereo} and the maximum depth integration range is set to \SI{10}{\meter}.

Due to the vast GPU memory requirements of RayFronts, quantitative benchmarking of the semantic accuracy was restricted to coarse voxel resolutions (\SI{0.5}{\meter}). While both approaches have similar runtimes, \FindAnything uses only $40\%$ of the memory thanks to the aggregation of \ac{VL} features at the segment level. As \supereight sometimes fails to preserve fine-grained geometric structures (e.g.\ poles) at such a coarse resolution, RayFronts achieves a higher class-mean recall. However, as most classes are coarser structures (e.g.\ roads, sidewalks), \FindAnything still yields a higher f-mIoU.
In contrast to RayFronts, which fails at finer resolutions (\SI{0.1}{\meter}) due to insufficient GPU memory, \FindAnything shows superior scalability and succeeds with a moderate memory usage at this resolution. The overhead in processing times is negligible, as our limitation is the image processing by the networks, while RayFronts shows an expected increase in runtime. Moreover, geometric details are better preserved at \SI{0.1}{\meter} resolution, resulting also in a higher semantic accuracy, significantly outperforming RayFronts$_{@0.5\mathrm{m}}$.

To assess the impact of the chosen segmentation model, we evaluate \FindAnything using \acs{eSAM} at half image resolution (FA - half-res) and replacing \acs{eSAM} with SAM2~\cite{ravi2024sam} (FA - SAM2).
While there is a big impact on runtime and memory usage, the semantic map accuracy stays largely unaffected. This is mainly because the different pixel accuracy of image segments is nullified when aggregating segments into the volumetric map for tracking. A higher map resolution is needed to benefit from more accurate segmentation.

\begin{figure}[t]
    \centering
    \includegraphics[width=0.95\columnwidth]{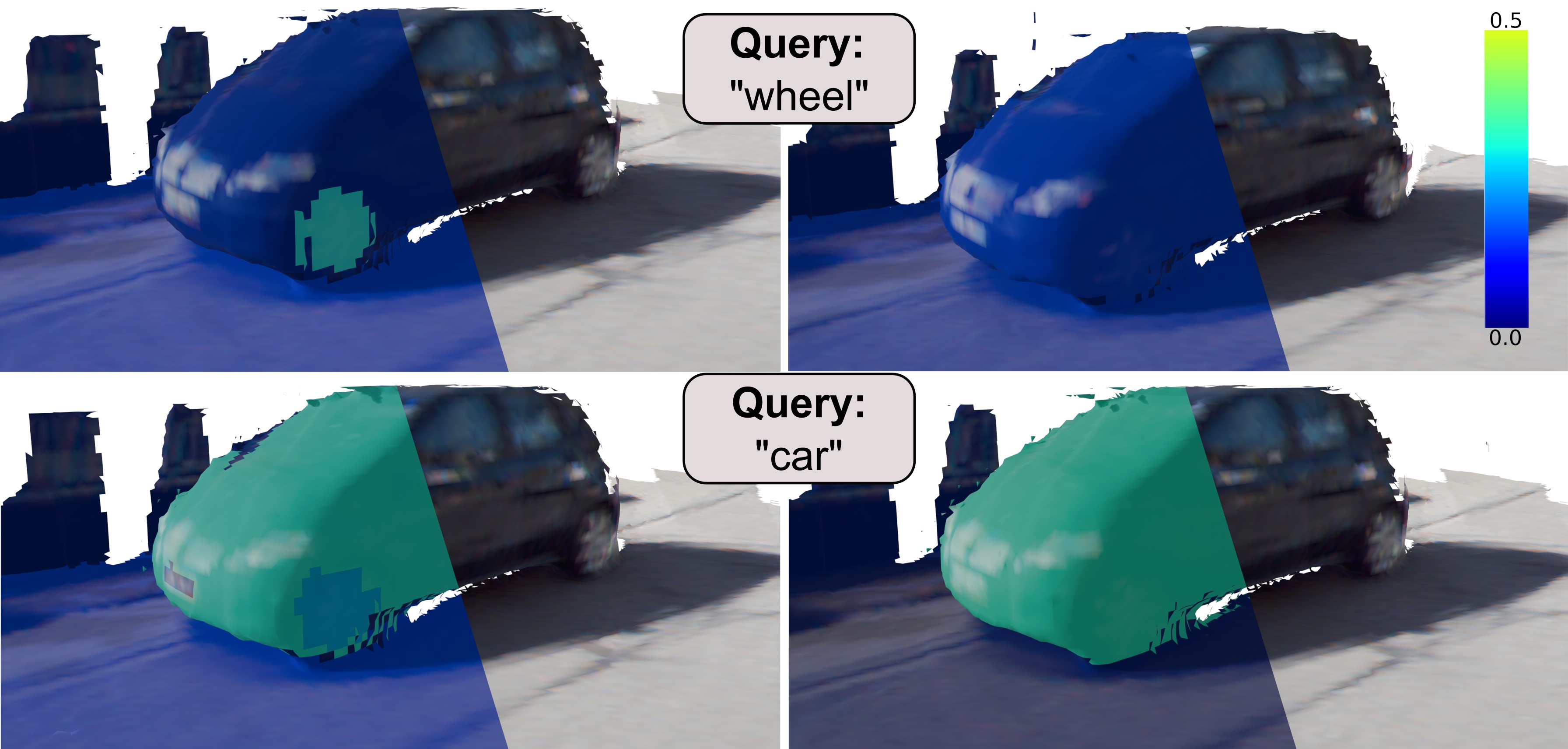}
    \caption{Qualitative example of the benefit of our oversegmentation approach. The left column shows the activations with our oversegmentation approach and the right column shows the activations for a method that never oversegments the objects in the map. The top row shows the activations for the query ``wheel'' and the bottom row for the query ``car''.}
    \label{fig:oversegmentation}
\end{figure}

Furthermore, the effectiveness of proposed feature fusion and oversegmentation strategy is demonstrated.
Table~\ref{tab:semantic-kitti} shows that the proposed weighted-mean feature embedding update (\cref{eq:feature_updates}) from different viewpoints improves semantic accuracy compared to storing the first feature vector obtained for each segment (FA - no fusion).
\Cref{fig:oversegmentation} qualitatively shows the benefit of the as-fine-as-possible segmentation strategy.
Oversegmentation allows distinguishing several parts of an object (e.g.\ ``wheel'') while maintaining an understanding of larger concepts or entities (e.g.\ ``car'').

These ablations show that the proposed configuration of \FindAnything offers the best balance between semantic accuracy, runtime, and memory usage.

Finally, we evaluate \FindAnything with SLAM instead of GT poses. We compute the $\mathrm{Sim(3)}$ alignment between the estimated and ground-truth trajectory and apply it to the maps before evaluation.
Semantic accuracy is degraded by drift, which is inevitable in such large-scale scenarios.
However, the comparison against \FindAnythingMonolithic proves the necessity for submaps under the presence of drift, as transforming them based on drift-correcting mechanisms (e.g.\ loop-closures) removes map inconsistencies.

\begin{table}
\caption{3D semantic map accuracy and resource usage (KITTI).} \label{tab:semantic-kitti}
\centering
\renewcommand{\arraystretch}{1.2} 
\begin{tabular}{c | l| c c| c| c }
    & & mAcc & f-mIoU & $T_{\mathrm{avg}}$[s] & $M_{\mathrm{avg}}$[GB] \\
    \hline
    \multirow{7}{*}{\rotatebox[origin=c]{90}{ \scalebox{0.9}{\scriptsize GT} }} & RayFronts$_{@0.5\mathrm{m}}$ & \textbf{45.85} & 35.60 & \textbf{277.9} & 24.61  \\
    & \FindAnything$_{@0.5\mathrm{m}}$ & 36.89 & \textbf{51.48} & 288.2 & \textbf{9.91} \\
    \cline{2-6}
    & RayFronts$^{1}_{@0.1\mathrm{m}}$ & - & - &  \textcolor{gray}{337.8}$^{2}$ & $>$24.5 \\
    & \textbf{\FindAnything (FA)} & \textbf{51.24} & \underline{53.90} & \underline{289.4} & \underline{16.23} \\
    \cdashline{2-6}
    & FA - SAM2 & 49.06 & \textbf{54.28} & 359.4 & 18.22 \\
    & FA - half-res & 50.59 & 53.29 & \textbf{124.5} & \textbf{15.11} \\
    & FA - no Fusion & \underline{49.88} & 50.26 & 286.3 & 16.24 \\
    \hline
    \multirow{2}{*}{\rotatebox[origin=c]{90}{\hspace{0.2cm}\scalebox{0.9}{\scriptsize SLAM} }} & \FindAnythingMonolithic & 23.81 & 28.24 & \textbf{285.7} & \textbf{15.58}\\
    & \FindAnything & \textbf{32.26} & \textbf{37.77} & 292.5 & 16.09\\ 
    \hline
\end{tabular}

\begin{tablenotes}
   \item $^{1}$ RayFronts$_{@0.1\mathrm{m}}$ failed on all sequences due to insufficient GPU memory (24.5 GB available).
   \item $^{2}$ RayFronts$_{@0.1\mathrm{m}}$ timings obtained by extrapolating the average processing time until the time of failure.
\end{tablenotes}
\end{table}

\subsection{Application: Autonomous \acs{MAV} Exploration}
\label{sec:evaluation-exploration}
To demonstrate the usefulness of our mapping approach to downstream tasks we integrate it into an autonomous \ac{MAV} exploration pipeline.
By using the map's \ac{VL} features to guide the exploration, we achieve high reconstruction accuracy of items or areas of interest that are selected online by natural language queries.
Thanks to our mapping method, the resulting exploration pipeline both works in 3D and can respond to on-the-fly queries, while other exploration approaches supporting open-vocabulary queries work only in 2D or require a priori known queries~\cite{wei2024ovexp,gadre2023cows,yokoyama2024vlfm}, making them unsuitable for 3D exploration.

We extend the sampling-based exploration planner for submaps proposed in~\cite{papatheodorou2025efficient} to account for our object-level language-based scene representation.
In~\cite{papatheodorou2025efficient}, $n_c \in \N^+$ candidate next views are sampled close to frontiers and ranked using an information-gain-over-time utility function.
The view with the highest utility is selected as the next goal and the process is repeated once the goal is reached.

The first extension over~\cite{papatheodorou2025efficient} consists of modifying the candidate next view sampling to take segments into account.
Each time a new goal view needs to be selected, the cosine similarity between the embedding $\mbf{f}_{\mathrm{q}}$ of the current natural language query and the language embedding $\mbf{\bar{f}}_{k}$ of each known segment is computed, selecting segments with a similarity greater than a predefined threshold $\beta \in \R^+$.
Each segment is associated with a \SI{1}{\cubic\meter} cube centered on the segment's volumetric centroid.
Since objects can be either over-segmented or represented in several submaps, a 3D non-maximum suppression is performed over the cubes, further reducing the number of segments considered.
Candidate next views are then sampled inside the segment cubes. Sampling inside the segment cubes stops once $n_c$ candidates have been accepted or if $3 n_c$ candidates have been sampled, with the remaining ones being sampled close to frontiers as in~\cite{papatheodorou2025efficient}.

The second modification of~\cite{papatheodorou2025efficient} is with regards to the utility function used to rank candidate next views.
The utility of each candidate $j$, as computed by~\cite{papatheodorou2025efficient}, is weighted with
\begin{equation}
    w_j = 
    \begin{dcases}
        \frac{\mbf{\bar{f}}_{{k}} \cdot \mbf{f}_{\mathrm{q}}}{\| \mbf{\bar{f}}_{{k}}\| \| \mbf{f}_{\mathrm{q}}\| } & \text{if } \frac{\mbf{\bar{f}}_{{k}} \cdot \mbf{f}_{\mathrm{q}}}{\| \mbf{\bar{f}}_{{k}}\| \| \mbf{f}_{\mathrm{q}}\| }\geq \beta\\
        \frac{\beta}{2}               & \text{otherwise},
    \end{dcases}
\end{equation}
where $\mbf{\bar{f}}_{{k}}$ is the \ac{VL} embedding of segment $k$ if the candidate is inside its cube or $\mbf{\bar{f}}_{{k}} = \mathbf{0}$ if the candidate is outside all segment cubes, and $\mbf{f}_{\mathrm{q}}$ is the \ac{VL} embedding of the query.

Quantitative results are obtained in the \texttt{00848-ziup5kvtCCR} scene of the Habitat-Matterport 3D dataset~\cite{hm3d}, using the semantic annotations from \cite{yadav2023habitat} to obtain ground-truth meshes.
The \ac{MAV} platform is simulated using Gazebo~\cite{gazebo}.
Exploration is evaluated using two natural language queries: the item of interest ``bed'' and the room ``bathroom''.
These queries were chosen because:
i)~the Habitat-Matterport 3D dataset~\cite{hm3d} contains only house scenes, limiting the potential queries;
ii)~they appear more than once in the scene;
iii)~they are contained in the ground-truth annotations of~\cite{yadav2023habitat} which do not include labels more closely related to \acs{SnR} scenarios;
and iv)~they are not visible from the \ac{MAV}'s starting location.
The baseline is the state-of-the-art submapping-based exploration method from~\cite{papatheodorou2025efficient} which lacks semantic information, demonstrating the benefit of incorporating open-vocabulary information.
Both \FindAnything and the baseline~\cite{papatheodorou2025efficient} use \SI{5}{\centi\meter} voxels.

\begin{figure}[tb]
    \centering
    \includegraphics[width=0.95\columnwidth]{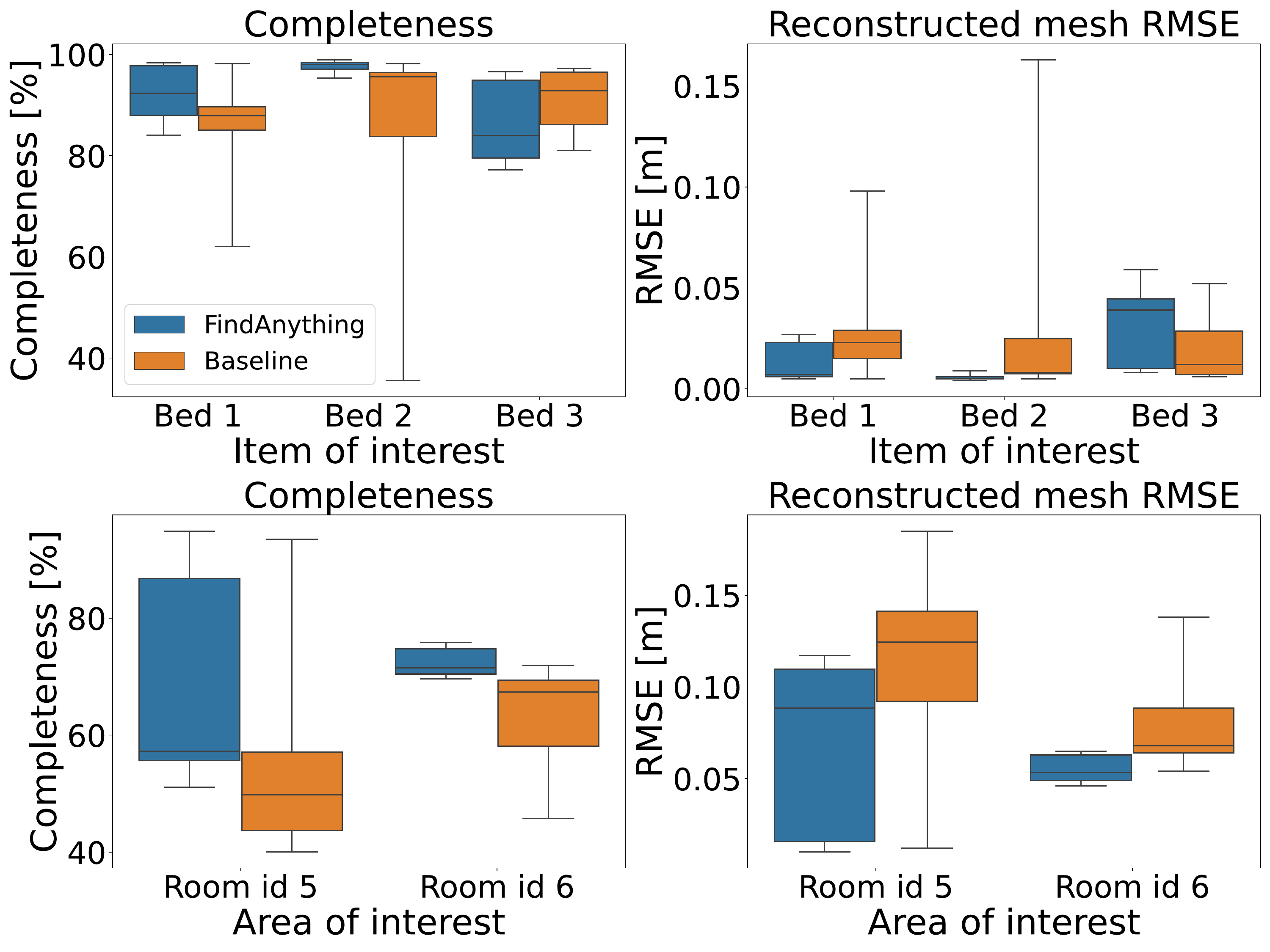}
    \caption{Mesh completeness (left column) and reconstruction \acs{RMSE} (right column) results of the 10th-90th percentiles for \FindAnything (blue) and baseline (orange) for
    queries ``bed'' (top row) and ``bathroom'' (bottom row).
    }
    \label{fig:find_anything_reconstruction}
\end{figure}

The mission is run 10 times for each combination of method and query, with each run lasting for 10 minutes, a typical \ac{MAV} flight time.
The final reconstructed map meshes are used to evaluate the completeness and the \ac{RMSE} against the ground-truth for each individual item or region of interest.
Completeness is computed as the percentage of ground-truth mesh vertices with a reconstructed mesh polygon within \SI{5}{\centi\meter}.
The results for the 10th to 90th percentiles are presented in \cref{fig:find_anything_reconstruction}.
\FindAnything achieves an overall higher completeness and mesh accuracy for both queries, while also being more consistent.
\FindAnything also yields a better reconstruction of the items of interest, with a larger margin for the area ``bathroom'' compared to the item ``bed''.
This is because a large part of the item is observable with even a single view, whereas areas of interest require continued observation which \FindAnything achieves by sampling candidate next views inside the segment cubes.

\begin{table}
\caption{Mean per-frame times for the stages of \FindAnything.}
\label{tab:replica-process-timings}
\centering
\setlength{\tabcolsep}{3.0pt}
\renewcommand{\arraystretch}{1.1} 
\begin{tabular}{l|lll}
\multirow{2}{*}{Stage} & \multicolumn{3}{c}{Mean time {[}ms{]}}                                             \\ \cline{2-4} 
                       & \multicolumn{1}{l|}{Replica} & \multicolumn{1}{l|}{Exploration Sim} & Exploration MAV \\ \hline
CLIP inference         & \multicolumn{1}{c|}{14} & \multicolumn{1}{c|}{14} & \multicolumn{1}{c}{340}\\
eSAM inference & \multicolumn{1}{c|}{105} & \multicolumn{1}{c|}{105}   & \multicolumn{1}{c}{171} \\ \hline
Render map segments & \multicolumn{1}{c|}{12} & \multicolumn{1}{c|}{7} & \multicolumn{1}{c}{16} \\ 
Segment tracking & \multicolumn{1}{c|}{109} & \multicolumn{1}{c|}{6} & \multicolumn{1}{c}{22} \\
\Acl{VL} fusion & \multicolumn{1}{c|}{30} & \multicolumn{1}{c|}{14} & \multicolumn{1}{c}{37} \\
Depth integration & \multicolumn{1}{c|}{40} & \multicolumn{1}{c|}{24} & \multicolumn{1}{c}{64} \\ \hline
\acs{SLAM} frontend & \multicolumn{1}{c|}{25} & \multicolumn{1}{c|}{41} & \multicolumn{1}{c}{32} \\
\acs{SLAM} backend & \multicolumn{1}{c|}{13} & \multicolumn{1}{c|}{24} & \multicolumn{1}{c}{16} \\
\end{tabular}

\begin{tablenotes}
   \item Horizontal lines separate stages that occur in different threads.
\end{tablenotes}
\end{table}

\subsection{Real-World Experiments}
\label{sec:evaluation-realworld}

To demonstrate \FindAnything's usefulness in potential \acs{SnR} scenarios, we simulated a firefighting use case by conducting an autonomous exploration experiment in an office environment. Throughout the mission, we defined target objects or areas relevant to fire response through natural language queries. We explored until a ``fire extinguisher'' was observed and then explored under the query ``kitchen'', as it is a house area with a higher risk of fire. \Cref{fig:real-world} shows the 3D reconstruction together with map activations for the queries ``fire extinguisher'' and ``exit''.
It shows that our system can explore the environment and build a volumetric representation suitable for navigation while aggregating the language embeddings in object-centric submaps.

We demonstrate that \FindAnything enables 3D exploration on resource-constrained devices by deploying it on a real-world \ac{MAV} relying only on onboard computation.
The system used in this experiment is a custom-built quadcopter (see~\cref{fig:system-overview}) equipped with a RealSense D455 stereo \mbox{RGB-D} camera and an NVIDIA Jetson Orin NX 16 GB computer.


To achieve faster \ac{eSAM} inference on the on-board computer, we downsample the image to half its original resolution and query 36 equally distributed points.
A timing break-down of the individual system components is presented in~\cref{tab:replica-process-timings}. To avoid sparse reconstructions due to slow network inference on the \ac{MAV}, we allow the integration of depth measurements also in the absence of CLIP features or \ac{eSAM} segments at a software-imposed rate of 3 Hz, to maintain the spirit of the original algorithm.

\section{Conclusion}
We have presented \FindAnything, a real-time open-vocabulary object-centric volumetric mapping framework, designed for scalability and online deployment on resource-constrained devices.
Using foundation models enables our system to be deployed in unknown environments without prior knowledge of the scene.
We have demonstrated \mbox{\FindAnything's} semantic mapping accuracy in both simulation and real-world indoor and outdoor environments, while being faster and significantly more memory efficient compared to baselines.
These capabilities make \FindAnything suitable for common robotics downstream tasks, such as exploration in \acs{SnR} scenarios and allows natural language-based interaction with the robot.
As a first of its kind, \FindAnything has been successfully deployed online on-board a resource-constrained \ac{MAV}.

In the future, we plan to push the boundaries for robotic exploration even further by incorporating semantic priors or hierarchical map representations to enable informed exploration even when lacking observations in the current scene. Furthermore, considering dynamic objects and people in the scene remains an open challenge to be addressed.

\bibliographystyle{IEEEtran}
\bibliography{IEEEabrv.bib, references.bib}

\end{document}